# A Face Recognition approach based on entropy estimate of the nonlinear DCT features in the Logarithm Domain together with Kernel Entropy Component Analysis

Arindam Kar[1],
[1] Indian Statistical Institute, Kolkata-700108, India
Email: kgparindamkar@gmail.com,

Debotosh Bhattacharjee[2], Dipak Kumar Basu[2*], Mita Nasipuri[2], Mahantapas Kundu[2]

[2]Department of Computer Science and Engineering, Jadavpur University, Kolkata- 700032, India
* Former Professor & AICTE Emeritus Fellow
Email: debotosh@indiatimes.com, dipakkbasu@gmail.com, mita_nasipuri@gmail.com, mkundu@cse.jdvu.ac.in

*Abstract*— This paper exploits the feature extraction capabilities of the discrete cosine transform (DCT) together with an illumination normalization approach in the logarithm domain that increase its robustness to variations in facial geometry and illumination. Secondly in the same domain the entropy measures are applied on the DCT coefficients so that maximum entropy preserving pixels can be extracted as the feature vector. Thus the informative features of a face can be extracted in a low dimensional space. Finally, the kernel entropy component analysis (KECA) with an extension of arc cosine kernels is applied on the extracted DCT coefficients that contribute most to the entropy estimate to obtain only those real kernel ECA eigenvectors that are associated with eigenvalues having high positive entropy contribution. The resulting system was successfully tested on real image sequences and is robust to significant partial occlusion and illumination changes, validated with the experiments on the FERET, AR, FRAV2D and ORL face databases. Experimental comparison is demonstrated to prove the superiority of the proposed approach in respect to recognition accuracy. Using specificity and sensitivity we find that the best is achieved when Renyi entropy is applied on the DCT coefficients. Extensive experimental comparison is demonstrated to prove the superiority of the proposed approach in respect to recognition accuracy. Moreover, the proposed approach is very simple, computationally fast and can be implemented in any real-time face recognition system.

*Index Terms*— Feature selection, discrete cosine transformation, entropy measure, entropy component analysis, Renyi Entropy Face recognition, Arc cosine kernel.

## I. Introduction

The discrete cosine transform (DCT) has the property that for a typical image, most of the visually significant information about the image is concentrated in just a few coefficients of the DCT. For this reason, the DCT is often used in image processing applications, and the features play a very important role in the task of recognition technique. Facial recognition technology, is one of the fastest growing fields in the biometric industry. A face recognition system aims to provide the correct label associated with that face from all the individuals in its database in case of occlusions, illumination changes and pose variations in face images. Basically, research in this field can be grouped into two main categories: holistic and feature based. Feature based methods extract information depending on the shapes and geometrical relationships of facial features, like eyes, mouth, nose and chin; whereas, holistic based methods extract global features using the high dimensional pixel intensity of an input facial image. There are many approaches to facial feature extraction techniques, for e.g. the eigen-face approach [1]; the linear discriminant analysis (LDA) [2]; the independent component analysis (ICA) [3, 4]; the Gabor wavelet analysis [5] etc. to 2D discrete cosine transform (DCT) [6]. As principal component analysis (PCA) derive features are sensitive to changes in the illumination direction [4], face recognition systems using PCA suffer rapid degradation in verification performance. All existing research on compressed domain is limited to DCT technique and DCT domain.

However, the facial images captured in uncontrolled and unconstrained environments still suffer a drop in the recognition performance with the existing face recognition techniques. The reason for this drop in the recognition performance is because of the appearance variations of the face image induced by various environmental factors, among which illumination is one of the most important factor. Changes in the illumination condition cause dramatic variation in face appearance and seriously affect the performance of face recognition systems. Hence, the effect of illumination variation is one of the challenging problems in a practical face recognition system. During the last several years, research on face recognition has been focused on reducing the impact of changes in lighting conditions, facial expression and poses [7].

To enhance the image classification information and improve the recognition effect, here we propose a new simple method based on the entropy measure (which includes the Shannon entropy [8], the Renyi entropy [9]) of the illumination normalized and jpeg quantized DCT coefficients per block over the whole entire image blocks to form the feature vector. Then these features for each image block are concatenated to construct a feature vector. Finally, kernel entropy component analysis (KECA) [10] with an arc cosine kernel extension is applied on these extracted feature vectors and the Mahalanobis distance [11] similarity measure is applied on the real KECA features for enhanced classification.

## II. Relevant Works

During the last several years, research on face recognition has been focused on reducing the impact of changes in lighting conditions, facial expression and poses. Among the several models proposed for face recognition, appearance based models gain much popularity because of their robustness against noise, occlusion and simplicity in terms of feature representation [12]. In these approaches, data transformation is a fundamental step and the goal is to obtain highly discriminative lower dimensional data from high-dimensional data. PCA and LDA are the widely used tools in face recognition domain that encode high-dimensional face images as lower dimensional Eigenfaces and Fisherfaces [13] respectively. However, the computational requirements of these approaches are greatly related to the dimensionality of the original data and the number of training samples. When the face database becomes larger, the time for training and the memory requirement will significantly increase. As a consequence, it is impractical to apply the PCA in systems with a large database. The discrete cosine transform (DCT) has been employed in face recognition [14], because it is a good approximation of principal component extraction, which helps to process and highlight the signal frequency features. The use of the DCT in face recognition becomes of more value than the PCA is firstly because of its computational speed, and secondly the DCT is data independent.

## III. Illumination Normalization in the Logarithm Domain

One ideal way of solving the illumination variation problem is to normalize a face image to a standard form under uniform lighting conditions. In the present work the normalization operation on the face image to remove the illumination variation problem has been done as follows;

i) by taking the logarithm of the image.

Let I(x, y) be the gray level distribution of an image. Thus the logarithmic transformation of the image is defined as:

$$if, I(x,y) = 0, I(x,y) = I(x,y) + \epsilon, \text{here } \epsilon = .0001$$
$$else, I(x,y) = log([I(x,y)])$$

So that the term log (0) can be compensated. ii) Then by variance equalization, the local contrast of the face in the image is increased, especially when the usable data of the image is represented by close contrast values. Through this adjustment, the intensity can be better distributed on the image histogram. This allows for areas of lower local contrast to gain a higher contrast without affecting the global contrast. The variance equalization is defined as a transformation on the input intensity levels ($i_k$) to obtain output intensity levels $v_k$ as :

$$v_k = \sum_{i=1}^{k} \frac{n_i}{\text{standard deviation of (I)}}, k = 1,2, \dots, L$$

where L = number of gray levels

Then the DCT transformation is used to transform the image from logarithmic domain to log DCT domain. Besides, it can be implemented using a fast algorithm which significantly reduces the computational complexity. The 2D DCT of an M × N image is defined as follows:

$$D(u,v) = a(u)a(v) \sum_{x=0}^{M-1} \sum_{y=0}^{N-1} I(x,y)$$
$$\times \cos\frac{(2x-1)u\pi}{2M} \cos\frac{(2y-1)v\pi}{2N}$$

(1)

and the inverse transform is defined as:

$$I(x,y) = \sum_{u=0}^{M-1} \sum_{v=0}^{N-1} a(u)a(v)D(u,v)$$
$$\times \cos\frac{(2x-1)u\pi}{2M} \cos\frac{(2y-1)v\pi}{2N}$$

(2)

where, u = 0,1, …, M − 1, v = 0,1, …, N − ; and a(u), a(v) are defined by :

$\alpha(u), \alpha(v) = \frac{1}{\sqrt{2}}$, for u, v = 0, and $\alpha(u), \alpha(v) = 1$, otherwise.

As the illumination variations are expected to be in the low-frequency components, the low-frequency components of a face image can be removed simply by setting the low-frequency DCT coefficients to zero. Evidently, the resulting system works like a high-pass filter. We can estimate the incident illumination on a face by using low-frequency DCT coefficients. It follows from (1) that setting the DCT coefficients to zero is equivalent to subtracting the product of the DCT basis image and the corresponding coefficient from the original image. If low-frequency DCT coefficients are set to zero, then we have

$$I'(x,y) = \sum_{u=0}^{M-1}\sum_{v=0}^{N-1} L(u,v) - \sum_{i=0}^{k} L(u_i, v_i)$$
$$= I(x,y) - \sum_{i=0}^{k} L(u_i, v_i) \quad (3)$$

$$L(u,v) = a(u)a(v)D(u,v)\sum_{x=0}^{M-1}\sum_{y=0}^{N-1} I(x,y)$$
$$\times \cos\frac{(2x-1)u\pi}{2M}\cos\frac{(2y-1)v\pi}{2N}$$
(4)

Further as the first DCT coefficient (i.e., the DC component) determines the overall illumination of a face image, the desired uniform illumination is be obtained by setting the DC coefficient to the mean of the overall log transformed image, i.e. $D(1,1) = \mu$, where $\mu = \sum_{i=1}^{M} \frac{D(u_i, v_i)}{M}$ = mean of the overall log transformed image. After normalizing the effect of the illumination inverse DCT, as shown in (2), is applied on the log DCT image to obtain the illumination normalized image in the logarithm domain.

### IV. Entropy-based Feature selection of the DCT transformed image in the logarithmic domain

Feature extraction is one of the most important steps in face recognition, and it usually attempts to reduce the high dimensional data space into the low dimensional feature vector. Dimension reduction of the feature vectors is essential for extracting the effective features and for reducing computational complexity in the classification step. But it is worthy to pay attention that even the most statistically dominant or discriminant features may degrade the system performance because of the existence of illumination or pose variations in face images. For instance, the relationship between first few principal components and illumination variations have been discussed in several papers [15,16].

The more significant facial features such as outline of hair and face, position of eyes, nose and mouth can be preserved by a very small number of low-frequency DCT coefficients. The results have shown that it is feasible and promising to extract discriminatory frequency components for classification. It is also worth paying attention that even the most dominant features may degrade the system performance because of the existence of variations on illumination or pose in face images. However, it is hard to determine which feature components are bounded with specific factors [17]. Once the jpeg quantization is performed on the DCT transformed blocks, we can use these DCT coefficients as a discretely sampled estimate of the image attribute probability density [18]. Here we use two types of entropy measures for the entropy estimates of each pixel containing maximum information of the face image for the final feature selection. The two entropy measures are used here:

(a) Boltzmann-Shannon Entropy: The Boltzmann-Shannon entropy [19] is first proposed as an information measurement in communication systems, and is defined as :

$$E = -\int p(x)\log(p(x))\,dx \quad (5)$$

(b) Renyi Entropy: The Renyi entropy [20, 21] is one of the canonical generalized entropies. It is defined as:

$$E = \frac{1}{1-q}\log\int p(x)^q\,dx, \quad (6)$$

where q is a variable called information order and p(x) is the probability density of the variable.

As the high-frequency components are related to unstable facial features such as expression and less stable than the low-frequency DCT coefficients. The steps for feature extraction are as follows:

Step 1 : The image $I(x,y)$ obtained from (2) is divided into blocks of size $G \times G$ pixels.

Step 2 : On each of these $G \times G$ blocks of image DCT is performed.

Step 3 : Then a compression is performed on each of the DCT block using the standard quantization matrix at quality factor 10, 50 and 90. Here the quality factor is chosen as 50.

Step 4: Then find the entropy estimate (E) of each pixel of the DCT transformed image of size $N \times N$ as follows;

i) For Shanon entropy measure, given in (5), the entropy estimate (E) is:

$$E = \sum_{i=1}^{N}\sum_{j=1}^{N}\left(p_{ij}\log(p_{ij}) + (1-p_{ij})\log(1-p_{ij})\right),$$

ii) For the Renyi entropy measure, given in (6), the entropy estimate (E) is :

$$E = \frac{1}{1-q}\sum_{i=1}^{N}\sum_{j=1}^{N}\log\left[p_{ij}^q + (1-p_{ij})^q\right]$$

where $p_i$ is the probability of the occurrence of the ith DCT coefficient. Thus $p_i$ consitutes the probability of occurrence of all the DCT coefficients defined as $p_i = dct(I(i,j))/(\sum_i \sum_j dct(I(i,j)))$.

Step 5: Select only that pixel which has the highest entropy estimate for the block.

Step 6: Store these entropy based feature points into the column vector $Y_i$.

Step 7: Finally concatenate each of these feature vectors to form a feature vector $Y$,

where $Y = Y_1, Y_2, ..., Y_N$, (7),

which contains the maximum amount of information of a whole face image of size $E * \left\lfloor \frac{M}{G} \right\rfloor \times \left\lfloor \frac{N}{G} \right\rfloor$, from all the Gabor convolution outputs. The magnitude of E is used as criterion for valid information selection. The magnitude of E indicates the held by a particular pixel in a particular position. The grater the value of E means the more amount of information content of the face image.

Step 8: The final high entropy content feature vector $Y_k$ is generated by accumulation of the elements of $Y_{I_{ij}}$ column wise to a single vector for the kth individual face.

In this work prior to classification, the dimensionalities of selected DCT coefficients, selected by virtue of the above mentioned entropy based criteria, are further reduced using the kernel entropy component analysis (KECA) [22]. This extracted entropy based feature vector $Y_{I_{ij}}$ is used as the input data instead of the whole image in KECA to derive the real KECA features $\mathcal{K}$.

## V. Kernel Entropy Component Analysis (KECA)

Applying the PCA technique to face recognition, Turk and Pentland [23] developed a well known Eigenfaces method, where the eigenfaces correspond to the eigenvectors associated with the largest eigenvalues of the face covariance matrix. The eigenfaces thus define a feature space, or "face space", which drastically reduces the dimensionality of the original space, and face detection and recognition are then carried out in the reduced space. Based on PCA, a lot of face recognition methods have been developed to improve classification accuracy and generalization performance [24, 25, 26]. The PCA technique, however, encodes only for second order statistics, namely the variances and the covariances. As these second order statistics provide only partial information on the statistics of both natural images and human faces, it might become necessary to incorporate higher order statistics as well. PCA is thus extended to a nonlinear form by mapping nonlinearly the input space to a feature space, where PCA is ultimately implemented. Due to the nonlinear mapping between the input space and the feature space, this form of PCA is nonlinear and naturally called nonlinear PCA [27]. Applying different mappings, nonlinear PCA can encode arbitrary higher-order correlations among the input variables. The nonlinear mapping between the input space and the feature space, with a possibly prohibitive computational cost, if implemented explicitly by kernel PCA [28]. The KECA data transformation method described below is fundamentally different from other spectral methods in two very important ways as follows: a) the data transformation reveals structure related to the Renyi entropy of the input space data set and b) this method does not necessarily use the top eigenvalues and eigenvectors of the kernel matrix. This new method is called as kernel entropy component analysis or simply as KECA.

The Renyi entropy estimator can be expressed as [20,21]:

$$H(p) = -\log \int p^2(x)dx, \qquad (8)$$

where, $p(x)$ is the probability density function. As logarithm is a monotonic function (5) can be formulated as $V(p) = \int p^2(x)dx = \epsilon_p(p)$, where $\epsilon_p(\ )$, denotes expectation with respect to x.

So, $H(p)$ can be estimated from $V(p)$ using a parzen window estimator, as [20,21]:

$$\hat{V}(p) = \frac{1}{N^2} \sum K(\chi, \chi_i), \qquad (9)$$

$\hat{V}(p)$ is the estimate of $V(p)$ where, K is the so called parzen window or kernel defined as $K = K(\mathcal{K}, \mathcal{K}_i)$ centered at $\mathcal{K}_t$ which computes an inner product in the Hilbert space F as: $K(\mathcal{K}, \mathcal{K}_i) = \Phi(\mathcal{K})^t \Phi(\mathcal{K}_t)$, where $\Phi$ is a non linear mapping between the input space and the feature space defined as: $\Phi: \mathbb{R}^N \to F$. Renyi estimator can thus be expressed in terms of the eigen values and vectors of the kernel matrix, as:

$$\hat{V}(p) = \frac{1}{N^2} \sum_{i=1}^{N} \left( \sqrt{\lambda_i} e_i^T \underline{1} \right)^2, \qquad (10)$$

where $\underline{1}$ is a $N \times 1$ vector with all elements equal to unity. The Renyi entropy estimator can further be eigen decomposed as $K = EDE^T$, where D is diagonal matrix storing the eigenvalues $(\lambda_1, \lambda_2, ..., \lambda_N)$ and E is a matrix with the corresponding eigen vector $e_1, ..., e_N$ as column and the matrix K is the familiar kernel matrix as used in KPCA. Each term in this expression will contribute to the entropy estimate. Thus using KPCA the projection $\Phi$ onto the ith principal axes $u_i$ is:

$$P_{u_i} \Phi = \sqrt{\lambda_i} e_i, \qquad (11)$$

From (10) it reveals that the Renyi entropy estimator is composed of projections onto all the kernel PCA axes. However, only a principal axis $u_i$ for which $\lambda_i > 0$ and $e_i^T 1 \neq 0$ contribute to the entropy estimate. Those principal axes contributing most to the Renyi entropy estimate clearly carry most of the information regarding the shape of the probability density function generating the input space data set. Let $\mathcal{K}$ be the extracted most informative feature vector, whose image in the feature space be $\Phi(\mathcal{K})$. The KECA feature E of $\mathcal{K}$ is derived as:

$$E = P_{u_i}\Phi(\mathcal{K}), \quad (12)$$

In order to derive real KECA features only those KECA eigenvectors are considered that are associated with nonzero positive eigenvalues and sum of the components of the corresponding eigenvectors not equals to zero.

## VI. KECA using Arc cosine kernels

Let $\chi_1, \chi_2, \ldots, \chi_n \in \mathbb{R}^N$ be the data in the input space, and $\Phi$ be a nonlinear mapping between the input space and the feature space, $\Phi: \mathbb{R}^N \longrightarrow F$. The nth arc cosine kernel [29] is defined by:

$$K_n(x,y) = 2 \int dw \frac{e^{-\left|\frac{w^2}{2}\right|}}{(2\pi)^{d/2}} \ominus (w.x) \quad (13)$$

$$*\ominus (w.y)(w.x)^n(w.y)^n,$$

The final result is most easily expressed in terms of the angle θ between the inputs x and y:

$$\theta = \cos^{-1}\left(\frac{x.y}{\|x\|\|y\|}\right) \quad (14)$$

For the general case, the nth order kernel function in this family can be written as:

$$K_n(x,y) = \frac{1}{n}\|x\|^n\|y\|^n J_n(\theta) \quad (15)$$

where all the angular dependence is captured by the functions $J_n(\theta)$. These functions are given by:

$$J_n(\theta) = (-1)^n(\sin\theta)^{2n+1}\left(\frac{1}{\sin\theta}\frac{\partial}{\partial\theta}\right)^n\left(\frac{\pi-\theta}{\sin\theta}\right) \quad (16)$$

As one important theoretical aspect of considering the arccosine kernels is that the arc cosine kernels of different degrees have qualitatively different geometric properties. In particular, for some kernels in this family, the surface in Hilbert space is described by a curved Riemannian manifold; for another kernel, this surface is flat, with zero intrinsic curvature; finally, for the simplest member of the family, this surface cannot be described as a manifold at all. It seems that the family of arc-cosine kernels exhibits a larger variety of behaviors than other popular families of kernels. Here, we used the arc-cosine kernels of degree n = 2.

The KECA produces a transformed data set with a distinct angular structure, in the sense that even the nonlinearly related input space data are distributed in different angular directions with respect to the origin of the kernel feature space. It also reveals cluster structure and reveals information about the underlying labels of the data. These obtained cluster results are used in image classification. Classification results obtained are comparable or even better in some cases than KPCA and other popular approaches [28].

## VII. Similarity Measures and Classification

The optimal features obtained from (12) are the representative of the images. The minimum distance between the optimal feature vectors of the images are used for image classification. The classifier then applies, the nearest neighbor (to the mean) rule for classification using the similarity (distance) measure δ as shown below:

$$\delta(\mathfrak{I}, M_k^o) = \min_j \delta(\mathfrak{I}, M_k^o) \rightarrow \mathfrak{I} \in w_k \quad (17)$$

The optimal feature vector $Y_{opt}$ is classified to that class of the closest mean $M_k'$ using the three different similarity measures. The similarity measures used here are, $L_2$ distance measure [30], $\delta_{L_2}$, cos the cosine similarity measure [30], $\delta_{cos}$, and the Mahalanobis distance measure $\delta_{Md}$, which are defined by:

$$\delta_{L_2} = (X-Y)^T(X-Y) \quad (18)$$

$$\delta_{cos} = \frac{-X^TY}{\|X\|\|Y\|} \quad (19)$$

$$\delta_{Md} = (X-Y)^T\Sigma^{-1}(X-Y) \quad (20)$$

Here Σ is the covariance matrix; ‖ ‖is the $L_2$ norm operator; and T is the transpose operator.

## VIII. Experimental Results and Analysis

Extensive experiments were carried out to illustrate the efficacy of the proposed approach. Essentially, four standard databases, i.e., FERET [31], AR [32], FRAV2D, [33], and ORL [34] have been addressed. These databases incorporate several deviations from the ideal conditions, including pose, illumination, occlusion, and gesture alterations. Several standard evaluation protocols reported in the face recognition literature have been adopted and a comprehensive comparison of the proposed approach with the state-of-the-art techniques has been presented. It is appropriate to indicate that the developed approach has been shown to perform well for the cases of severe gesture variations and contiguous occlusion with little change in pose, scale, illumination, and rotation. However, it is not meant to be robust to other deviations such as severe pose and illumination variations. From the experiments we learn that the Renyi estimator performs better than the Shanon estimator. So Renyi estimator has been used throughout the work for enhanced face recognition purpose.

### 8.1 The ORL Face Database

The ORL database [34], is maintained at the AT&T Laboratories, Cambridge University. It consists of 40 subjects with 10 images per subject. The database incorporates facial gestures, such as smiling or non smiling, open or closed eyes, and alterations like glasses or without glasses. It also characterizes a maximum of 20 degree rotation of the face with some scale variations of about 10 percent. Figure 1 shows all samples of an individual from the ORL database.

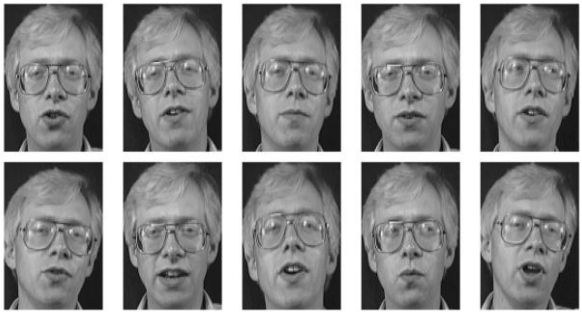

Figure 1. A typical individual from the ORL database

We follow two evaluation protocols as proposed in the literature quite often [35,36,37,38]. Evaluation Protocol 1 (EP1) takes the first five images of each individual as a training set, while the last five are designated as probes. But here we further restrict the EP1 by considering only the first two images of each individual as training set, while the last eight are designated as probes. For Evaluation Protocol 2 (EP2), the "leave-one-out" strategy is adopted. All experiments are conducted by down-sampling $112 \times 92$ images to an order of $10 \times 5$. A detailed comparison of the results for the two evaluation protocols is summarized in table I. For EP1, the entropy based DCT approach achieves a comparable recognition accuracy of 99 percent in a 20 dimensional feature space; the best results are reported for the latest Eigen feature Regularization and Extraction (ERE) approach [38], which are 2.0% poorer better than the proposed method using a 50 dimensional feature space.

Table I: Experimental results for EP1 and EP2 using the ORL Database.

| Evaluation Protocol | Methods | Accuracy (%) |
|---|---|---|
| EP1 | PCA | 80.5% |
| | ICA | 85% |
| | Fisher Faces | 94.5% |
| | Kernel Eigen Faces | 94% |
| | 2DPCA | 96% |
| | ERE | 97% |
| | **Proposed method** | 99% |
| EP2 | PCA | 88.5% |
| | ICA | 93.8% |
| | Fisher Faces | 97.5% |
| | Kernel Eigen Faces | 98% |
| | 2DPCA | 98..3% |
| | ERE | 99.25 |
| | **Proposed method** | 99.5% |

## 8.2 The FRAV2D face database

The FRAV2D face database [33], employed in the experiment consists of 1100 colour face images of 100 individuals, 11 images of each individual are taken, including frontal views of faces with different facial expressions, under different lighting conditions. All colour images are transformed into gray images and scaled to $92 \times 112$. Figure 2 shows all samples of one individual. The details of the images are as follows: (A) regular facial status; (B) and (C) are images with a 15° turn with respect to the camera axis; (D) and (E) are images with a 30° turn with respect to the camera axis; (F) and (G) are images with gestures; (H) and (I) are images with occluded face features; (J) and (K) are images with change of illumination.

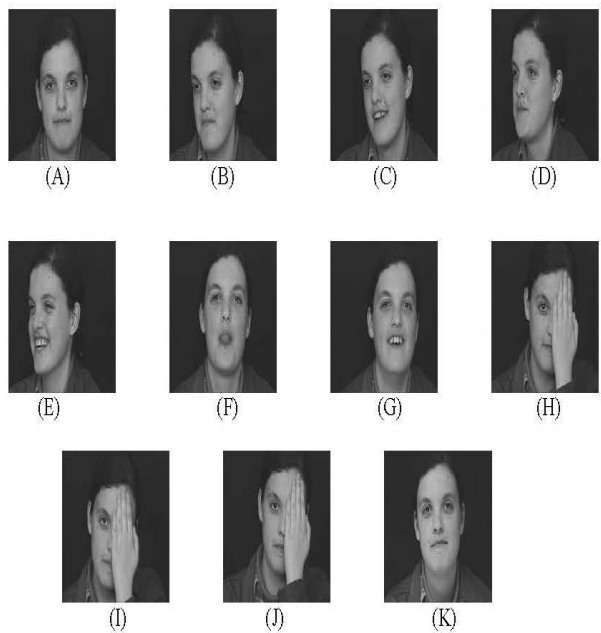

Figure 2 Demonstration images of an individual from the FRAV2D database

We follow an evaluation protocol as proposed in the literature quite often [35, 36, 37, 38]. Evaluation protocol 1 (EP1) takes the first 3 images (A-C), of a particular individual are selected as training samples from the 11 images of the particular individual, and the remaining 8 images of that individual are used as testing samples. Average recognition results of the proposed method in comparison with other linear subspace approaches like PCA, ICA, KECA, on the FRAV2D database is shown in table II.

Table II: Experimental results on the FRAV2D database using the EP1.

| Method | Recognition (%) Using different no. of training samples | | Average Recognition Rates (%) |
|---|---|---|---|
| | 3 | 4 | |
| PCA | 80 | 85 | 82.5 |
| ICA | 83.7 | 89 | 86.35 |
| DCT (only) | 81.5 | 84.75 | 83.25 |
| KECA (only) | 85 | 90.75 | 87.875 |
| GWT | 85.5 | 89.5 | 87.5 |
| GWT-LDA | 88.3 | 90.33 | 89.33 |
| **Proposed Method** | **96.125** | **97.75** | **96.9** |

## 8.3 The AR Face Database

The AR database consists of more than 4,000 color images of 126 subjects (70 men and 56 women) [32], and characterizes divergence from ideal conditions by incorporating various facial expressions (neutral, smile, anger, and scream), luminance alterations (left light on, right light on, and all side lights on), and occlusion modes (sunglass and scarf). It has been used by researchers as a testbed to evaluate and benchmark face recognition algorithms. The color images in AR database [32] are converted to gray scale and cropped into the size of $120 \times 70$, same as the image size used in [39, 40]. There were 50 subjects with 12 images of frontal illumination per subject used in [39], and the same amount of subjects with 14 non-occluded images per subject were used in [41]. In our experiment, 75 subjects with 14 non-occluded images per subject are selected from the AR database. The first seven images of all subjects are used in the training, and the remaining seven images serve as testing images. Some normalized face images are shown in Figure 3.

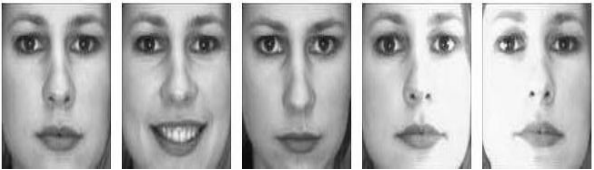

Figure 3. Normalized face images from the AR database.

Table III: shows face recognition results of different methods on the AR images with different gestures. From the results it is revealed that for images with expression changes, on the AR subset the entropy based DCT approach together with KECA included the arc cosine kernels achieves better average face recognition rates than the other linear subspace approaches like PCA, ICA, Edgemap, linear regression (LRC) classification[42] .

Table IV : For images with lighting changes from the AR database, the entropy based DCT approach together with KECA included the arc cosine kernels achieves better average face recognition rates than the other linear subspace approaches like PCA, ICA, Edgemap, linear regression (LRC) classification[42]. This is coincide with the results reported in [43,44].

Table III: Face recognition results of different methods on the AR database images with different gestures.

| Methods | Different Gestures | | | | Over all Accuracy (%) |
|---|---|---|---|---|---|
| | Neutral | Smile | Anger | Scream | |
| PCA | 90.98 | 80.45 | 42.86 | 71.43 | 68.3 |
| ICA | 98.5 | 80.7 | 44.0 | 75.6 | 74.7 |
| DCT | 90 | 79 | 50 | 68 | 71.75 |
| KECA | 93 | 80 | 78 | 80 | 82.75 |
| Edge Map | 96.99 | 95.49 | 48.12 | 78.45 | 87.72 |
| Linear Regression | 99.0 | 98.5 | 98.5 | 99.5 | 98.88 |
| **Proposed Method** | **99** | **99** | **98** | **99** | **98.75** |

Table IV: Face recognition results of different methods on the AR database images with different lighting conditions.

| Methods | Different Lighting conditions | | | Average Accuracy (%) |
|---|---|---|---|---|
| | left light on | right light on | all side lights on | |
| PCA | 66.17 | 51.88 | 77.44 | 65.16 |
| ICA | 70 | 55 | 77.5 | 67.5 |
| DCT | 88.6 | 88.5 | 85 | 87.36 |
| KECA | 93.4 | 94 | 95 | 94.13 |
| Edge Map | 98.50 | 98.50 | 93.98 | 96.99 |
| Linear Regression | 100 | 100 | 83.27 | 94.43 |
| **Proposed Method** | **99** | **99** | **98** | **98.67** |

## 8.4 The FERET database

### 8.4.1. FERET- Evaluation Protocol 1 (EP1):

The FERET database is arguably one of the largest publicly available databases [31]. Following [38, 45], we construct a subset of the database consisting of 128 subjects, with at least four images per subject. We, however, used four images per subject [38]. Figure 4 shows images of a typical subject from the FERET database. It has to be noted that, in [38], the database consists of 256 subjects; 128 subjects (i.e., 512 images) are used to develop the face space, while the remaining 128 subjects are used for the face recognition trials. The proposed entropy based DCT approach uses the gallery images of each person to form a linear subspace; therefore, it does not require any additional development of the face space. However, it requires multiple gallery images for a reliable construction of linear subspaces. Cross-validation experiments for entropy based approach were conducted in a 42 dimensional feature space; for each recognition trial, three images per person were used for training, while the system was tested for the fourth one. The results are shown in Table V. The frontal images fa and fb incorporate gesture variations with small pose, scale, and rotation changes, whereas ql and qr correspond to major pose variations (see [46], for details). The proposed approach copes well with the problem of facial expressions in the presence of small pose variations, achieving high recognition rates of 98.25 and 98.75 percent for fa and fb, respectively. It outperforms the benchmark PCA and ICA algorithms by margins of 22.5 and 22.75 percent for fa and 21.5 and 26.75 percent for fb, respectively. The proposed approach, however, shows degraded recognition rates of 90.45 and 85 percent for the severe pose variations of ql and qr, respectively; however, even with such major posture changes, it is substantially superior to the PCA, ICA, direct DCT, LRC and the KECA approaches. In an overall sense, we achieve a recognition accuracy of 93.44 percent, which is favorably comparable to 84.00 percent recognition achieved by ERE [38] using single gallery images.

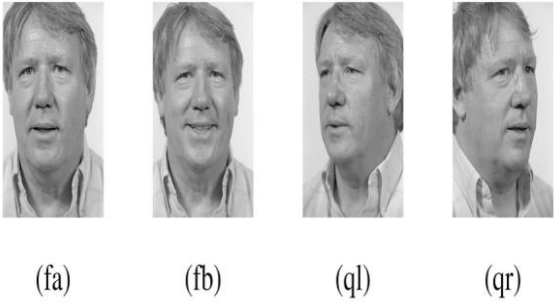

Figure 4 A typical subject from the FERET database; fa and fb represent frontal shots with gesture variations, while ql and qr correspond to pose variations.

**8.4.2. FERET- Evaluation Protocol 2 (EP2):**

To validate the robustness of the proposed entropy based technique on DCT transformed image in the logarithm domain together with KECA, with an arc cosine kernel extension approach, the FERET pose database [47] has been used which consists of images from 200 subjects. For each subject, images have been captured at viewpoints ba ; bi ; bh ; bg ; bf ; be ; bd ; bc and bb which correspond to viewpoint angles of $0°$ frontal ; $-60°$ pose angle (bi); $-40°$ pose angle (bh); $-25°$ pose angle (bg); $-15°$ pose angle (bf); $+15°$ pose angle (be); $+25°$ pose angle (bd); $+45°$ pose angle (bc) ; $+60°$ pose angle (bb). As the images include the background and the body chest region, so in the experiments each image is cropped to exclude those, transformed into gray images and is scaled to $\times n$ ($92 \times 112$). Figure 5 shows all samples of one subject. The details of the images are as follows: (A) $0°$ frontal face (i.e., ba); (B) $+15°$ pose angle (i.e., be); (C) $-15°$ pose angle (i.e., bf); (D) $+25°$ pose angle (i.e., bd); (E) $-25°$ pose angle (i.e., bg); (F) $+40°$ pose angle (i.e., bc); (G) $-40°$ pose angle (i.e., bh); (H) $+60°$ pose angle (i.e., bb); (I) $-60°$ pose angle (i.e., bi); (J) alternative expression ((i.e., bj); (K) different illumination (i.e., bk).

The EP2 is restricted by considering only the firstly 3 images shown in figure 5, A (ba), B (be) and C (bf) of a particular individual, used as training samples and the remaining eight images of the particular individual are used for testing samples. The same process is repeated by considering on the first 2 images that is A (ba), B (be) as training images and the remaining nine images for testing. The pose FERET database is intentionally used in order to evaluate the effectiveness of combination of entropy based DCT features together with KECA in presence of pose variation.

We calculate the recognition rates in two cases: firstly using only the frontal face image (ba) and only the $\pm 15°$ images as training images (be, bf) and checked across for various pose variations like bi, bh, bg, bd, bc, bb and expression, illumination changes for bj and bk respectively. Secondly we used only the frontal image (ba) and the $+15°$ pose variation image as training images and similarly and checked across for various pose variation like bi, bh, bg, bd, bc, bb and expression, illumination changes for bj and bk respectively. For all case studies, the proposed entropy based DCT approach is found to be superior to the benchmark, Gabor wavelet transformation (GWT), Local Binary Patterns (LBP) [48], PCA and ICA approaches, and to the most recent works on the FERET dataset.

The proposed approach showed quite agreeable results with the large database as well. It persistently achieved high recognition rates of 98.25 and 98.50 percent, for fb and fc, respectively. For the case of severe pose variations of bi and bb, we note a slight degradation in the performance, as expected. Using the three training images (ba,be,bf) the average recognition rate achieved is 95.75% and using the two training images (ba,be) the average recognition rate achieved is 92.5%. The overall performance is, however, pretty much comparable with an average recognition success of 94.13 percent. For all case studies, the proposed entropy based DCT approach is found to be superior to the benchmark, Gabor wavelet transformation (GWT), PCA and ICA approaches, and to the most recent works on the FERET dataset.

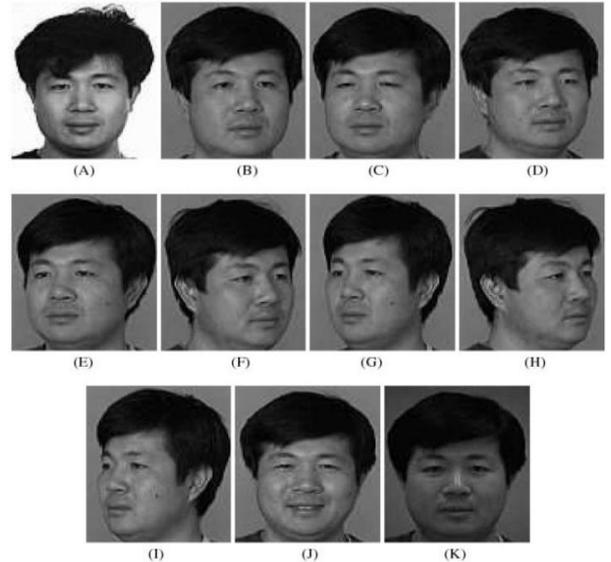

Figure 5 Demonstration images of an individual from the pose FERET database.

| Method | FERET Probe Sets | | | | |
| --- | --- | --- | --- | --- | --- |
| | Fb | Fc | ql | qr | Overall |
| PCA | 74.5 | 73.5 | 66 | 73 | 71.75 |
| ICA | 75 | 72 | 66 | 74 | 71.75 |
| GWT | 96 | 97 | 79 | 81 | 88.25 |
| DCT(only) | 88 | 89.5 | 70.43 | 72.6 | 80.13 |
| KECA (only) | 89.43 | 88.63 | 88 | 82 | 87.02 |
| Phillips et. al. | 96 | 82 | 62 | 60 | 75.00 |
| LRC[42] | 92 | 94.5 | 98.13 | 84.38 | 92.25 |
| Grassmann registration manifolds [49] | 98 | 98 | 80 | 84 | 90.00 |
| LGBP [50] | 98 | 97 | 74 | 71 | 85.00 |
| **Proposed method using RBF kernels** | **98** | **98** | **89** | **85** | **92.5** |
| **Proposed method using arc cosine kernels** | **98.25** | **98.5** | **90** | **87** | **93.44** |

Table VI: Recognition results of different methods on the pose FERET sets using evaluation protocol 2 (EP2):

| Method | Pose FERET Sets | | |
|---|---|---|---|
| | Number of Training Samples | | |
| | 3 | 2 | |
| | Using ba, be and bf as training images | Using ba, be as training images | Overall |
| PCA | 75.67 | 74.75 | 75.21 |
| ICA | 76.475 | 73.67 | 75.07 |
| GWT | 79.65 | 82.5 | 81.08 |
| DCT(only) | 89.5 | 70.43 | 79.97 |
| KECA (only) | 89.43 | 88.63 | 89.03 |
| LBP | 90.67 | 92.77 | 91.72 |
| LRC[42] | 94.5 | 92 | 93.25 |
| LGBP [23] | 93 | 90 | 91.50 |
| **Proposed method using RBF kernels** | **89.67** | **85.93** | **87.8** |
| **Proposed method using arc cosine kernels** | **95.75** | **92.5** | **94.13** |

## IX. Conclusion

This paper presents a novel three-step feature selection algorithm taking the advantages of discrete cosine transform together with entropy measures and entropy component analysis. The first step of proposed method is to reduce images to comparable low-dimension frequency components using DCT. Then, entropy estimates are used to search optimal combination of DCT coefficients guided by entropy measures function. The third part is to further reduce dimensionality using the kernel entropy component analysis with an arc cosine kernel approach been discussed. Experimental results show that the evolutionary selected features perform better than original DCT coefficients, and outperform most of existing approaches on benchmark face databases. Hence the proposed technique can be implemented in real time face recognition system. In our future work we will focus on the elimination/reduction of shadowing effect.

## Acknowledgement

Authors are thankful to a major project entitled "Design and Development of Facial Thermogram Technology for Biometric Security System," funded by University Grants Commission (UGC), India and "DST-PURSE Programme" and CMATER and SRUVM project at Department of Computer Science and Engineering, Jadavpur University, India for providing necessary infrastructure to conduct experiments relating to this work.

## References

[1] P. Belhumeur, J. Hespanha, and D. Kriegman, "Eigenfaces vs. Fisherfaces: Recognition using class Specific linear projection," *IEEE Trans. on PAMI*, vol.19, no.7, 1997.

[2] H. Yu and J. Yang, "A direct LDA algorithm for high dimensional data with application to face recognition". Pattern Recognition, 34(10), 2001, pp. 2067-2070.

[3] P. Comon, "Independent component analysis, a new concept?" Signal Proc., vol. 36, pp. 287–314, 1994.

[4] Hyvärinen, A. and Oja, E. (1997). A fast fixed point algorithm for independent component analysis. Computation, 9(7):1483–1492.

[5] D. Gabor, "Theory of communication," J. IEE, vol. 93, pp. 429-459, 1946.

[6] Zhang Yan-Kun and Liu Chong-Qin, "A Novel Face Recognition Method Based on Linear Discriminant Analysis". J. Infrared Millim.Wav. 22(5), 2003, pp. 327-330.

[7] X. Zou, J. Kittler and K. Messer, "Illumination Invariant Face Recognition: A Survey", in: Proceedings of 1st IEEE International Conference on Biometric: Theory, Application and Systems (2007), pp. 1–8.

[8] C. E. Shannon. The mathematical theory of communication. Bell System Technical Journal, pages 379 –423, 623–653, July and October 1948.

[9] A. Renyi. Some fundamental questions of information theory. Turan [13] (Originally: MTA III. Oszt. Kozl., 10, 1960, pp. 251-282), pages 526– 552, 1960.

[10] Robert Jenssen "Kernel entropy component analysis" IEEE transactions on pattern analysis and machine intellegence vol. 32, no. 5, May 2010.

[11] http://en.wikipedia.org/wiki/Mahalanobis_distance .

[12] Hyung-Soo Lee and Daijin Kim, "Illumination-Robust Face Recognition Using Tensor-Based Active Appearance Model" 8[th] International Conference on Automatic Face & Gesture Recognition, 2008. FG '08.

[13] K. Etemad and R. Chellappa, "Discriminant analysis for recognition of human face images", Journal of the Optical Society of America A, Vol. 14, No. 8, pp. 1724-1733, 1997.

[14] Z. M. Hafed and M. D. Levine, "Face Recognition using the Discrete Cosine Transform". International Journal of Computer Vision, 43(3), 2001, pp. 167~188.

[15] A. S. Georghiades, D.J. Kriegman, and P. N. Belhumeur, "Illumination cones for recognition under variable lighting: Faces," in Proceedings of IEEE Conference on Computer Vision and Pattern Recognition, 1998, pp. 52–58.

[16] Z. H. Sun, G. Bebis, and R. Miller, "Object detection using feature subset selection," Pattern Recognition, vol. 37, pp. 2165–2176, 2004.

[17] Huan Liu, Edward R. Dougherty, Feature-Selection Overfitting with Small-Sample Classifier Design, IEEE Intell. Sys. Nov-Dec 2005.

[18] A. Nobuhide and T. Masaru. Information theoretic learning with maximizing Tsallis entropy. In Proceedings of International Technical Conference on Circuits/System, Computers and Communications, ITC-CSCC, pg. 810–813, Phuket, Thailand, 2002.

[19] ] C. E. Shannon. The mathematical theory of communication. Bell System Technical Journal, pages 379 –423, 623–653, July and October 1948.

[20] A. Renyi, "On Measures of Entropy and Information," Selected Papers of Alfred Renyi, vol. 2, pp. 565-580, Akademiai Kiado, 1976.

[21] http://mathworld.wolfram.com/RenyiEntropy.html.

[22] Robert Jenssen "kernel entropy component analysis" IEEE transactions on pattern analysis and machine intellegence vol. 32, no. 5, May 2010.

[23] M. Turk and A. Pentland, "Eigenfaces for recognition," Journal of Cognitive Neurosicence, vol. 3, no. 1, pp. 71-86, Mar. 1991.

[24] Wayo Puyati, Walairacht, A. "Efficiency Improvement for Unconstrained Face Recognition by Weightening Probability Values of Modular PCA and Wavelet PCA" in the 10th Int. Conference on Advanced Communication Technology, 2008. ICACT 2008.

[25] M.H. Yang, N. Ahuja, and D. Kriegman, "Face Recognition Using Kernel Eigenfaces," Proc. IEEE Int'l Conf. Image Processing, Sept. 2000.

[26] Jian Yang; Zhang, D.; Jing-Yu Yang "Constructing PCA Baseline Algorithms to Reevaluate ICA-Based Face-Recognition Performance", in IEEE Trans. on Systems, Man, and Cybernetics, Part B, Vol. 37, Issue4, 2007.

[27] Xudong Xie; Kin-Man Lam "Gabor-based kernel PCA with doubly nonlinear mapping for face recognition with a single face image", IEEE Transactions on Image Processing, Vol. 15 , Issue: 9 ,2006.

[28] Chengjun Liu "Gabor-based kernel PCA with fractional power polynomial models for face recognition" in IEEE Transactions on Pattern Analysis and Machine Intelligence, Vol. 26 , Issue: 5 ,2004.

[29] Youngmin Cho, Lawrence K. Saul, "Analysis and Extension of Arc-Cosine Kernels for Large Margin Classification," Neural Computation, 22(10):2678–2697, (2010).

[30] http://reference.wolfram.com/mathematica/guide/DistanceAndSimilarityMeasures.html.

[31] The Face recognition technology (FERET) face database, http://www.itl.nist.gov/iad/humanid/feret.

[32] A.M. Martinez and R. Benavente, "The AR Face Database," CVC Technical Report #24, June 1998.

[33] Face Recognition and Artificial Vision group FRAV2D face database http://www.frav.es/.

[34] Olivetti & Oracle Research Laboratory, The Olivetti & Oracle Research Laboratory Our Database of Faces, http://www.cam-orl.co.uk/facedatabase.html.

[35] J. Yang, D. Zhang, A.F. Frangi, and J. Yang, "Two-Dimensional PCA: A New Approach to Appearance-Based Face Representation and Recognition," IEEE Trans. Pattern Analysis and Machine Intelligence, vol. 26, no. 1, pp. 131-137, Jan. 2004.

[36] M.H. Yang, "Kernel Eignefaces vs. Kernel Fisherfaces: Face Recognition Using Kernel Methods," Proc. Fifth IEEE Int'l Conf. Automatic Face and Gesture Recognition, pp. 215-220, May 2002.

[37] P.C. Yuen and J.H. Lai, "Face Representation Using Independent Component Analysis," Pattern Recognition, vol. 35, no. 6, pp. 1247-1257, 2002.

[38] X. Jiang, B. Mandal, and A. Kot, "Eigenfeature Regularization and Extraction in Face Recognition," IEEE Trans. Pattern Analysis and Machine Intelligence, vol. 30, no. 3, pp. 383-394, Mar. 2008.

[39] A.M. Martinez, "Recognizing Imprecisely Localized, Partially Occluded, and Expression Variant Faces from a Single Sample per Class," IEEE Trans. Pattern Analysis and Machine Intelligence, vol. 24, no. 6, pp. 748-763, June 2002.

[40] B.G. Park, K.M. Lee, and S.U. Lee, "Face Recognition Using Face- ARG Matching," IEEE Trans. Pattern Analysis and Machine Intelligence, vol. 27, no. 12, pp. 1982-1988, Dec. 2005.

[41] H. Cevikalp, M. Neamtu, M. Wilkes, and A. Barkana, "Discriminative Common Vectors for Face Recognition," IEEE Trans. Pattern Analysis and Machine Intelligence, vol. 27, no. 1, pp. 4-13, Jan. 2005.

[42] Imran Naseem,Roberto Togneri, Mohammed Bennamoun, "Linear Regression for Face Recognition", IEEE Transaction on Pattern Analysis and Machine Intelligence, Vol. 32, no. 11, November 2010.

[43] Y. Gao, M.K.H.Leung, "Face recognition using line edge map," IEEE Transactions on Pattern Analysis and Machine Intelligence, vol. 24, no. 6, pp. 764-779, 2002.

[44] J. Song, B. Chen, and Z. Chi, X. Qiu, and W. Wang, "Face recognition based on binary template matching," In Proceedings of the Third International Conference on Intelligent Computing (ICIC2007), Springer Publishers, Berlin/Heidelberg, pp. 1131-1139, August, 2007.

[45] J. Lu, K.N. Plataniotis, A.N. Venetsanopoulos, and S.Z. Li, "Ensemble- Based Discriminant Learning with Boosting for Face Recognition," IEEE Trans. Neural Networks, vol. 17, no. 1, pp. 166-178, Jan. 2006.

[46] J. Phillips, H. Wechsler, J. S. Huang, and P. J. Rauss, The FERET database and evaluation procedure for face recognition algorithms. Image and Vision Computing, 16(5):295-306, 1998.

[47] P.J. Phillips, H. Moon, S. A. Rizvi, and P. J. Rauss, "The FERET evaluation methodology for face-recognition algorithms," IEEE Trans. Pattern Anal. Mach. Intell., vol. 22, no. 10, pp. 1090–1104, Oct. 2000.

[48] Ngoc-Son Vu, Alice Caplier, "Efficient Statistical Face Recognition across pose using Local Binary Patterns and Gabor wavelets," IEEE 3rd International Conference on Biometrics: Theory, Applications, and Systems, 2009. BTAS '09. .

[49] Lui, Y.M. and Beveridge, J.R. "Grassmann registration manifolds for face recognition". In "Proceedings of the European Conference on Computer Vision", pages 44–57 (2008).

[50] Zhang, W., Shan, S., Gao, W., Chen, X. and Zhang, H. "Local gabor binary pattern histogram sequence (lgbphs): A novel non-statistical model for face representation and recognition". In "Proceedings of the IEEE International Conference on Computer Vision", (2005).

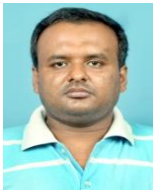

**Arindam Kar** received his M.Sc and M.Tech. degrees from Indian Institute of Technology (Kharagpur), India in 1998 and 2000 respectively. At present he is persuing Ph. D. from Jadavpur University, India, and working as an Associate Scientist in Indian Statistical Institute, Kolkata, India. His areas of interest includes Computer Vision, Image Processing, Genetic Algorithms, and Watermarking Techniques.

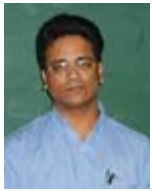

**Debotosh Bhattacharjee** received the MCSE and Ph. D.(Eng.) degrees from Jadavpur University, India, in 1997 and 2004 respectively. He was associated with different institutes in various capacities until March 2007. After that he joined his Alma Mater, Jadavpur University. His research interests pertain to the applications of computational intelligence techniques like Fuzzy logic, Artificial Neural Network, Genetic Algorithm, Rough Set Theory, Cellular Automata etc. in Face Recognition, OCR, and Information Security. He is a life member of Indian Society for Technical Education (ISTE, New Delhi), Indian Unit for Pattern Recognition and Artificial Intelligence (IUPRAI), and member of IEEE (USA).

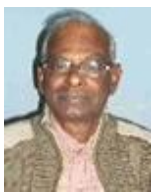

**Dipak Kumar Basu** received his B.E.Tel.E., M.E.Tel., and Ph.D. (Engg.) degrees from Jadavpur University, in 1964, 1966 and 1969 respectively. Prof. Basu has been a faculty member of J.U from 1968 to January 2008. He is presently an A.I.C.T.E. Emiretus Fellow at the CSE Department of J.U. His current fields of research interest include pattern recognition, image processing, and multimedia systems. He is a senior member of the IEEE, U.S.A., Fellow of I.E. (India) and W.B.A.S.T., Kolkata, India and a former Fellow, Alexander von Humboldt Foundation, Germany.

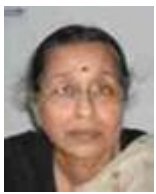

**Mita Nasipuri** received her B.E.Tel.E., M.E.Tel.E., and Ph.D. (Engg.) degrees from Jadavpur University, in 1979, 1981 and 1990, respectively. Prof. Nasipuri has been a faculty member of J.U since 1987. Her current research interest includes image processing, pattern recognition, and multimedia systems. She is a senior member of the IEEE, U.S.A., Fellow of I.E (India) and W.B.A.S.T, Kolkata, India.

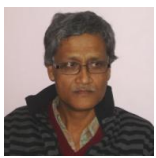

**Mahantapas Kundu** received his B.E.E, M.E.Tel.E and Ph.D. (Engg.) degrees from Jadavpur University, in 1983, 1985 and 1995, respectively. Prof. Kundu has been a faculty member of J.U since 1988. His areas of current research interest include pattern recognition, image processing, multimedia database, and artificial intelligence.